\documentclass{article}

\usepackage{microtype}
\usepackage{graphicx}
\usepackage{subfigure}
\usepackage{booktabs} 

\usepackage{hyperref}

\usepackage[accepted]{icml2024}

\usepackage{amsmath}
\usepackage{amssymb}
\usepackage{mathtools}
\usepackage{amsthm}

\usepackage[capitalize,noabbrev]{cleveref}

\theoremstyle{plain}

\theoremstyle{definition}

\theoremstyle{remark}

\usepackage{listings}

\usepackage[textsize=tiny]{todonotes}

\icmltitlerunning{Frequency-Time Diffusion with Neural Cellular Automata}

\begin{document}

\twocolumn[
\icmltitle{Frequency-Time Diffusion with Neural Cellular Automata}



\icmlsetsymbol{equal}{*}

\begin{icmlauthorlist}
\icmlauthor{John Kalkhof}{yyy}
\icmlauthor{Arlene Kühn}{yyy,sch}
\icmlauthor{Yannik Frisch}{yyy}
\icmlauthor{Anirban Mukhopadhyay}{yyy}
\end{icmlauthorlist}

\icmlaffiliation{yyy}{Darmstadt University of Technology, Karolinenplatz 5, 64289 Darmstadt, Germany}
\icmlaffiliation{sch}{German Cancer Research Center (DKFZ), Heidelberg, Germany}

\icmlcorrespondingauthor{John Kalkhof}{john.kalkhof@gris.tu-darmstadt.de}

\icmlkeywords{Machine Learning, ICML}

\vskip 0.3in
]



\printAffiliationsAndNotice{} 

\begin{abstract}
Despite considerable success, large Denoising Diffusion Models (DDMs) with UNet backbone pose practical challenges, particularly on limited hardware and in processing gigapixel images. To address these limitations, we introduce two Neural Cellular Automata (NCA)-based DDMs: Diff-NCA and FourierDiff-NCA. Capitalizing on the local communication capabilities of NCA, Diff-NCA significantly reduces the parameter counts of NCA-based DDMs. Integrating Fourier-based diffusion enables global communication early in the diffusion process. This feature is particularly valuable in synthesizing complex images with important global features, such as the CelebA dataset. We demonstrate that even a 331k parameter Diff-NCA can generate $512 \times 512$ pathology slices, while FourierDiff-NCA (1.1m parameters) reaches a three times lower FID score of 43.86, compared to the four times bigger UNet (3.94m parameters) with a score of 128.2. Additionally, FourierDiff-NCA can perform diverse tasks such as super-resolution, out-of-distribution image synthesis, and inpainting without explicit training.

\end{abstract}

\begin{figure*}[htbp]
 \centering
  {\includegraphics[width=.98\linewidth]{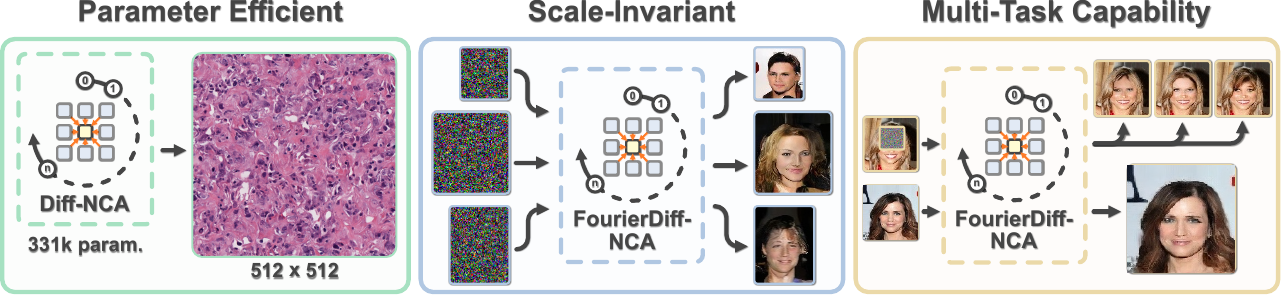}}
  \caption{Diff-NCA is parameter efficient while being able to generate infinite seamless images. The one-cell model size allows FourierDiff-NCA to be applied to inputs, different from the training size, thus generating images of different shapes and scales. This same architecture also allows it to efficiently regenerate parts of an image in an inpainting task and perform superresolution on an existing image, without the need for retraining.}
  \label{fig:Figure1}
\end{figure*}

\section{Introduction}
Denoising Diffusion Models (DDMs) have emerged as the leading architecture for generating high-quality images \cite{dhariwal2021diffusion}. However, UNet-based \cite{ronneberger2015u} DDMs have significant limitations, such as a model size of tens to hundreds of millions of parameters, that restrict their use in environments with limited hardware. But even when powerful hardware is available, the inability to adapt to different image sizes and the inefficiency when increasing the generative size makes synthesizing gigapixel images, such as digital pathology scans or satellite images, challenging.

Neural Cellular Automata (NCAs) \cite{mordvintsev2020growing, gilpin2019cellular} offer a promising architectural alternative to the UNet. Unlike traditional deep learning models, NCAs adopt a unique single-cell model architecture inspired by biological cell communication. They interact only with their immediate neighbors, keeping the model size small while efficiently encoding information. Local communication makes NCAs highly adaptable to different image sizes \cite{mordvintsev2021texture, pajouheshgar2023dynca}.
However, initial efforts at NCA-based image generation encountered limitations. Synthesizing beyond image sizes of $64\times64$ faced obstacles due to the inherent need for local communication, which requires increasing steps for global knowledge communication. Increasing the step count not only slows down the run-time but also complicates the learning process and increases the required VRAM during training time \cite{kalkhof2023med}.

We introduce two new DDM methods based on Neural Cellular Automata to address these challenges and jumpstart NCA-based DDMs: \textbf{Diff-NCA and FourierDiff-NCA}.
Diff-NCA focuses on local features of the underlying distribution, making it suitable for applications where local details are crucial, such as digital pathology scans or satellite imagery. With merely 331k parameters, it can synthesize digital pathology scans, multiple magnitudes larger than the training size of $64\times64$, which we demonstrate by synthetizing images of size $512 \times 512$.
We introduce FourierDiff-NCA to recognize the importance of global features in datasets such as CelebA \cite{liu2015faceattributes}. It uses a Fourier-based diffusion approach to connect the frequency-structured Fourier space to the image space by starting diffusion in the Fourier domain and completing it in the image space. This method solves global communication and simplifies the optimization path using a quarter of the Fourier space. FourierDiff-NCA can generate $64 \times 64$ CelebA images with merely 1.1m parameters while maintaining global knowledge.

In our evaluation, we compare our proposed FourierDiff-NCA with UNet-based DDMs and VNCA \cite{palm2022variational} on the CelebA dataset. FourierDiff-NCA achieves an FID score of 43.86 with merely 1.1m parameters, outperforming four times bigger UNets with 3.94m parameters and an FID score of 128.2 and the ten times bigger VNCA with an FID score of 299.9. Moreover, we demonstrate our architecture's flexibility by highlighting its ability to perform super-resolution, inpainting, and synthesize images of out-of-distribution sizes, further supporting the potential of our proposed models for various image-generation tasks.
Diff-NCA and FourierDiff-NCA represent a considerable advance in NCA-based image generation, jumpstarting NCA-based DDMs. They offer scalable, efficient, and adaptable solutions (illustrated in Figure \ref{fig:Figure1}) that overcome the limitations of traditional DDMs and provide a path to high-quality image synthesis at arbitrary scales on minimal hardware.

Upon acceptance, we will make our complete framework available under \url{github.com/anonymized}.

\section{Related Work}
While Neural Cellular Automata can generalize across different image sizes, current NCA image generation approaches, such as VNCA \cite{palm2022variational} and GANCA \cite{otte2021generative}, have not exploited this property. These methods are limited to $64 \times 64$ pixel images and show constraints in their random sampling results. This unexplored potential motivates our work with Diff-NCA and FourierDiff-NCA.

\subsection{Neural Cellular Automata (NCA)}
NCAs differ substantially from conventional deep learning architectures as they are built on a one-cell model (a detailed introduction to NCAs can be found in the appendix \ref{app:NCAIntro}). This one-cell model is replicated in each cell of an image, where it communicates only with its direct neighbors. To gain global knowledge, this learned update rule is repeated multiple times, which means that a naive NCA requires at least 100 steps to communicate across a $100 \times 100$ image. Due to the size of the single-cell model, NCAs typically require below a million parameters to handle complex tasks such as growing an image from a single cell \cite{mordvintsev2020growing} or image generation \cite{otte2021generative, palm2022variational}. Although the model sizes are small, the VRAM requirements for NCAs increase exponentially during training as backpropagation requires all duplications of the NCA in the world and across all timesteps. This makes it challenging to train NCAs on a large number of steps \cite{kalkhof2023med}.

\subsection{NCA Image Generation}
Current NCA image generation methods are based on two main architectures: variational autoencoders (VAEs) \cite{kingma2013auto} and generative adversarial networks (GANs) \cite{goodfellow2014generative}.

\textbf{GAN-based:} GANCA \cite{otte2021generative} explores the generation of emojis with a size of $30 \times 30$ pixels, using an NCA as a generator in a traditional GAN framework augmented by a classical discriminator.

\textbf{VAE-based:} VNCA \cite{palm2022variational} expands images up to $64 \times 64$ pixels by integrating an NCA into the image generation component of a VAE \cite{kingma2013auto}. By distributing the latent vector of the VAE to each cell of the NCA, the method can regenerate the encoded image.

While these two methods show the general capability of synthesizing images using NCAs, they face the challenge of global communication within the image. The authors of VNCA approach this by starting the generation with a reduced image scale and then gradually doubling it. Although this strategy increases communication speed within the image, the need for increasing duplications as the image scale grows makes it challenging to learn meaningful update rules. In addition, communication across the image still requires multiple iterative steps, which limits efficiency.

\begin{figure*}[tp]
 \centering
  {\includegraphics[width=.98\linewidth]{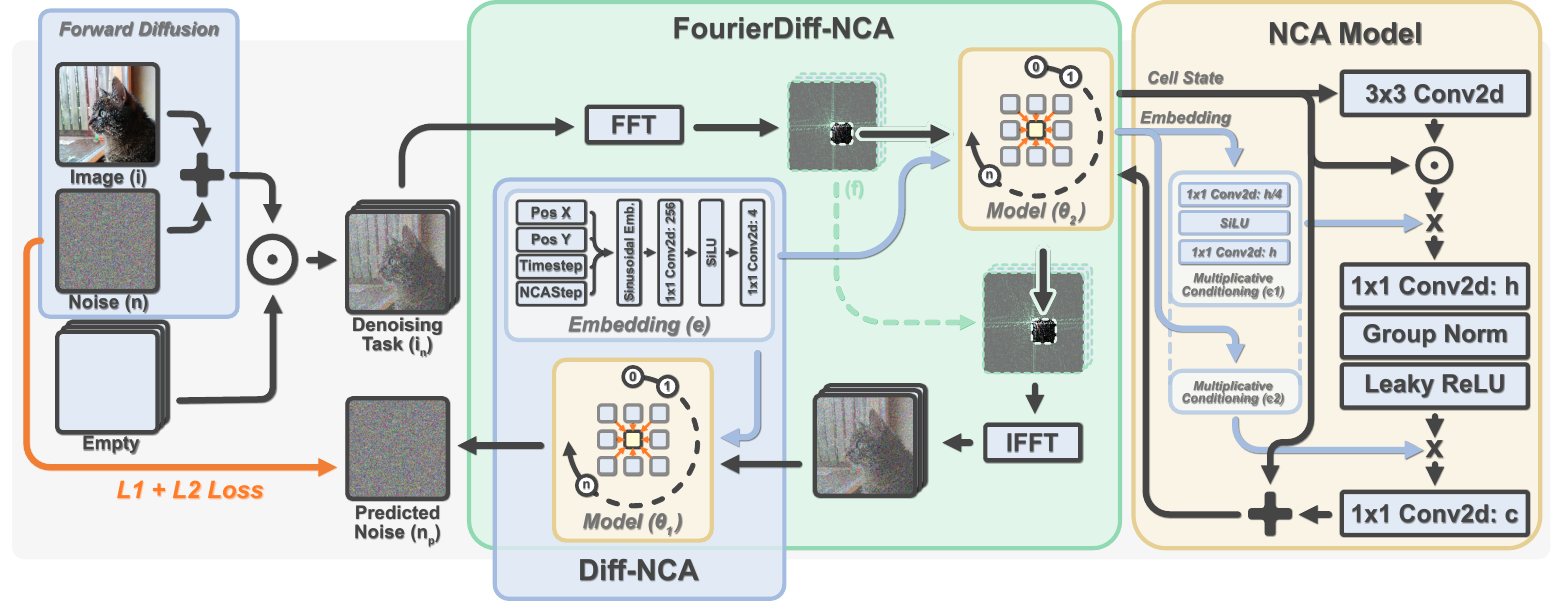}}
  \caption{Diff-NCA predicts the noise using iterative local communication of NCA's, whereas FourierDiff-NCA additionally utilizes the Fourier space to communicate global knowledge across the image space.}
  \label{fig:Diff_NCA}
\end{figure*}

\subsection{Denoising Diffusion Models}
Denoising Diffusion Models \cite{ho2020denoising} are based on the idea of sequentially diffusing an unknown distribution into a univariate Gaussian \cite{sohl2015deep} distribution. Then, using adequately small diffusion steps, one can train a denoising model to inverse these steps. Eventually, new samples from the unknown target distribution can be generated by sampling from the known distribution and repeatedly applying the denoising model, sequentially pushing the sample back into the target distribution. Several incremental improvements have been proposed over the recent years \cite{song2020denoising, nichol2021improved, rombach2022high, hoogeboom2023simple, croitoru2023diffusion, gao2023masked}, and this family of models has shown superior image generation quality compared to the GAN-based standard to date \cite{dhariwal2021diffusion}.

Typically, the denoising model is a modified version of a UNet \cite{ronneberger2015u}, which comes with several limitations: UNets are typically compute-heavy architectures with several millions of parameters. Further, their performance is bound to the training modalities and can not out-of-the-box produce images of varying scales or be used for tasks such as inpainting or superresolution. For such tasks, adapted architectures are required for training UNet-based DDMs \cite{rombach2022high,ho2022cascaded,saharia2022image,li2022sdm,lugmayr2022repaint}. On the contrary, our proposed architecture is flexible for different input modalities, such as different scales, and it can be used out-of-the-box for tasks like inpainting and superresolution.

\section{Methodology}
Neural Cellular Automata are one-cell models that iteratively approach a final goal through a learned update rule. Building on this basis, we present Diff-NCA. This image-generation methodology combines the diffusion process with the generalization capability of NCAs to generate images of varying sizes. Moreover, we introduce FourierDiff-NCA to address the local communication limitations of NCAs, which are significant obstacles in capturing global knowledge. FourierDiff-NCA circumvents this limitation using the Fourier space and bridges distant parts of an image without multiple iterations.

\subsection{Fourier Space: Single-Step Global Communication}

When using Neural Cellular Automata, one of the major limitations is the number of steps required for the model to acquire global knowledge. Since NCAs communicate with their direct neighbors, communication across a $100\times100$ image in a naive setup thus requires 100 steps. The inherent structure of the Fourier representation (we explain this in-depth in the appendix \ref{app:fourierCom}), where lower frequency data lives in the middle of a two-dimensional Fourier space, 
enables a fundamental shift in the communication pattern. This shift allows NCAs operating in this space to achieve global \emph{communication across the entire image in a single step}, a stark contrast to the linear step-wise progression needed in image space. 
Additional iterations can be used to refine the transmitted signal. The possibility of instantaneous communication over the entire image space is a significant advantage, that arises from applying an inverse Fast Fourier Transform following the initial NCA communication, whereby all relevant data in the limited Fourier space is transferred to the global scale. After the initial acquisition of global information in Fourier space, each cell in the image space starts from a comprehensive understanding of the global context and then adjusts the details based on local information, which is a clear departure from the traditional "detail-first" approach.

Notably, only a fraction of the Fourier space is required for this process since a significant portion of the information in the Fourier space is assumed to be insignificant from a quality point of view (we further illustrate this in the appendix in Figure \ref{fig:fft_reasoning}). 

\subsection{FourierDiff-NCA Architecture}
In FourierDiff-NCA, we address the challenge of achieving global coherence by beginning the diffusion process in Fourier space to capture global information and then transitioning to image space to integrate local details. This approach allows us to combine global and local information effectively without the need for a high number of NCA steps. We use the separate NCA models $m_1$ and $m_2$ for the image and Fourier space. The denoising process is illustrated in Figure \ref{fig:Diff_NCA}.

\textbf{Diff-NCA: } As a subset of the FourierDiff-NCA architecture, Diff-NCA uses only local communication (illustrated in the appendix in Figure \ref{fig:perceptiveRange}). It is noteworthy that Diff-NCA can be run independently of FourierDiff-NCA, resulting in a diffusion process based solely on the image space and thus considering only local features. The denoising task $i_n$ follows the procedure of DDMs and serves as the input of Diff-NCA. It is a combination of the input image $i$ and the noise $n$. An embedding includes the position $x, y$ of the NCA, the timestep $t$ of diffusion, and the NCA timestep. Diff-NCA predicts the noise $n_p$ from $i_n$ by iterating $m_1$, $s$ times over the image, incrementing its perceptive range by 1 per step. The loss is computed between $n_p$ and $n$ using a combined L2 and L1 loss, which diverges from the original L2 loss, to enhance convergence.

\textbf{Embedding:} The embedding's linear information of pos x, y, diffusion timestep, and NCA timestep is processed using sinusoidal encoding \cite{vaswani2017attention}, as introduced by DDM \cite{ho2020denoising}. A sequence of a linear layer of size 256, SiLU activation, and another linear layer mapping to four output channels $e$ is utilized and concatenated with the input. Subsequently, this encoding is then multiplied with the output of the first $3 \times 3$ convolutional layer of Diff-NCA as well as the output of the second $1 \times 1$ convolutional layer. As we use multiplicative conditioning, we use two additional multiplicative embedding blocks \(M_{eb1}\), \(M_{eb2}\), that map the four output channels to the required sizes of $2h$ and $h$ respectively, built of a $1 \times 1$ convolution, another SiLU, and a second $1 \times 1$ convolution. 




\textbf{FourierDiff-NCA:} extends Diff-NCA by initiating the diffusion process by gathering global information in Fourier space. Through a Fast Fourier Transform (FFT) on $i_n$, FourierDiff-NCA receives the diffusion task $f$ in the Fourier space. We extract a $16 \times 16$ cell window starting at the center of $f$ paired with the embedding $e$, simplifying communication. This $16\times16$ quarter of the Fourier space, contains enough details for global communication. After 32 iterations (required to communicate once across and back of a $16\times16$ world) of $m_2$ in Fourier space, an inverse FFT is performed to convert back to the image space, and the process transitions to Diff-NCA, providing initial global information in the channels $c$ of each NCA cell.


\subsection{Model Architecture}
We design our model, illustrated in Figure \ref{fig:Diff_NCA}, with simplicity in mind. We keep the architecture identical for the image and Fourier space. However, there is a difference in the number of input channels $c$, since converting data to the Fourier space results in two values per channel.

Given an input image $i$ in RGB format, the model is defined with three channels for input $I$, three channels for predicted noise $N$, and 90 additional empty channels $E$ for storing information between steps. The empty channels are essential in any NCA as they are the only medium for information retention between steps. Therefore, the total number of channels $c$ is given by: $c = I + N + E = 96$, containing the image, output noise, and the NCA's internal state $v$.

Local communication is implemented through a $3 \times 3$ 2D convolution. The output of that convolution is concatenation with the previous internal state $v$ and the embedding $e$. This concatenated vector is multiplied with the output of \(M_{eb1}\). The next layer is a $1 \times 1$ 2D convolution that maps the concatenated vector to a hidden vector of depth $h = 512$. Group normalization and a leaky ReLU activation are applied to ensure normalization and introduce nonlinearity. 

Now the output of the ReLU is multiplied by the output of \(M_{eb2}\). Afterwards, another $1 \times 1$ 2D convolution maps the hidden layer back to the channel size $c$, resulting in an output vector $o$. The internal state $v$ is updated by: $v \leftarrow v + o$, controlled by a random reset mechanism referred to as the fire rate, which is set to a probability of 90\%. When the reset mechanism is activated, the update of the cells in question is set to 0.

Following practices aligned with the leading methods in the field \cite{dhariwal2021diffusion, ho2022cascaded, Meng_2023_CVPR}, our model incorporates an exponential moving average (EMA) on the weights, with a decay rate of 0.99.

\begin{figure}[htbp]
 \centering
  {\includegraphics[width=.98\linewidth]{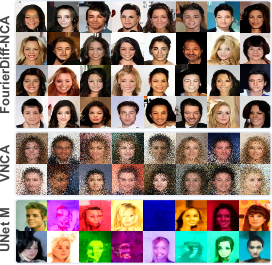}}
  \caption{Qualitative comparison between FourierDiff-NCA (1.85m), VNCA, and DDM based on UNet M. With a parameter count of 1.85m, 9.73m, and 3.94m, respectively. }
  \label{fig:GeneratedImages}
\end{figure}

\begin{figure*}[htbp!]
 \centering
  {\includegraphics[width=.98\linewidth]{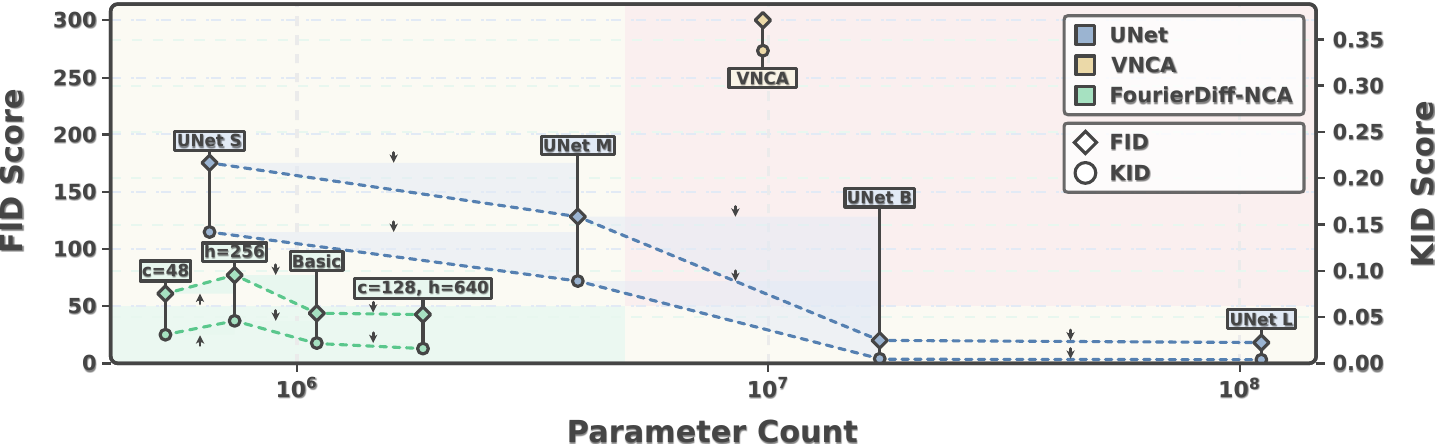}}
  \caption{Influence of parameter count to image generation performance for FourierDiff-NCA, UNet, and VNCA (the detailed numbers can be found in the appendix in Table \ref{tab:quant}). \emph{Green quadrant: low parameters, high performance; red quadrant: high parameters, low performance; yellow quadrants: tradeoff.}}
  \label{fig:quant}
\end{figure*}

\section{Experimental Results}
In the experimental results, our focus goes beyond FourierDiff-NCA's basic image generation capabilities. We are particularly interested in the unique properties that distinguish them from conventional models such as the UNet. We begin the evaluation with a comprehensive image quality comparison to highlight the difference between FourierDiff-NCA, conventional UNet-based architectures, and VNCA. We then test the suitability of FourierDiff-NCA in generating images at Out-Of-Distribution (OOD) scales, thus investigating its adaptability at unfamiliar scales. In addition, we address the aspect of efficient image inpainting by using pixel-precise activation of the model. This method enables targeted cell updates that improve efficiency while ignoring unaffected cells. In this context, we are also investigating the model's ability to supersample existing images and add additional details. Lastly, we investigate Diff-NCA's capability of generating seamless megapixel images.

\subsection{Data and Infrastructure}
For the evaluation of our proposed methods, we select two distinct datasets that present different challenges.

\textbf{CelebA dataset}: The CelebA dataset \cite{liu2015faceattributes}, a widely used benchmark, consists of 202,599 images, each of size $178 \times 218$. We scale all images to a uniform size of $64 \times 64$ to match the dataset with the input requirements of the UNet and VNCA. The data is split 80\%:10\%:10\% for training, validation, and testing. CelebA presents a challenge as it includes various facial images against different backgrounds. The inclusion of individuals with a range of accessories, hairstyles, and facial expressions further increases the complexity of the dataset. The size and variety of this dataset allow us to evaluate the ability of our models to handle intricate details and different visual features.

\textbf{BCSS Pathology dataset}: This dataset \cite{amgad2019structured} contains 144 high-resolution pathology samples, which present a unique challenge for generative models. Initially, some images contain blur at the lowest level and are resized by a factor of four in the x and y direction to obtain clear and concise visual patterns. We then extract patches of size $64 \times 64$ for training purposes. The BCSS pathology dataset provides an opportunity to rigorously test the capabilities of our proposed methods in generating large images in the medical domain. For this dataset, we use an 80\%:10\%:10\% split for training, validation, and testing. 

\textbf{Infrastructure}: All models are implemented in PyTorch \cite{paszke2019pytorch}, where we use the official implementation of VNCA \cite{palm2022variational}. The models are trained on an Ubuntu 22.04 system using an Nvidia RTX 3090 and an Intel i7-12700 processor Additional details can be found in the appendix and the codebase that we will make available upon acceptance.

\subsection{Metrics}
To assess the quality of the synthesized outputs, we use the well-established metrics Fréchet Inception Distance (FID) \cite{heusel2017gans} and Kernel Inception Distance (KID) \cite{binkowski2018demystifying}, which measure the similarity between the real and synthetic images based on the Inception-v3 \cite{szegedy2015going} model. For both evaluations, we use a set of 2048 real images from the test split compared to an equal number of synthesized images. 

\subsection{Qualitative Comparison: Image Synthesis}
Examining the images, FourierDiff-NCA (1.85m) generated in Figure \ref{fig:GeneratedImages} shows that the model learns global information. Interestingly, FourierDiff-NCA does not have enough steps in the image space to achieve global knowledge transfer. Nevertheless, it exhibits such behavior, indicating that communication occurs in the Fourier space. Thanks to this Fourier communication, the model can capture various features of the underlying distribution, such as variations in facial features, hair color, clothing, and facial expressions, all while using a comparably small parameter count of 1.85m. The efficiency becomes apparent when comparing it to the four times bigger VNCA and the 2.5 times bigger UNet M-based DDM, where the results look worse with blurry results and substantial color shifts. Although the results of FourierDiff-NCA are imperfect, they represent a promising step toward detailed image generation with Neural Cellular Automata and demonstrate the potential for complex generative tasks.

\subsection{Quantitative Comparison: Image Synthesis}
Figure \ref{fig:quant} compares FourierDiff-NCA, various UNet configurations, and VNCA. Notably, FourierDiff-NCA excels at the tradeoff between parameters and FID reaching an FID of 43.86 and KID of 0.022 with only 1.1m parameters, outperforming four times larger UNet-based DDMs with an FID of 128.2 and KID of 0.089 and the ten times bigger VNCA with an FID of 299.99 and KID of 0.338. 
Exploring scalability, we train an enlarged FourierDiff-NCA variant with $c=128$ and $h=640$, dubbed FourierDiff-NCA (1.85m), 68\% larger than the original. This model achieved a modest FID improvement at 42.69, but significantly enhanced the KID score to 0.016, indicating promising potential for future scaling of this method. This is further supported by the qualitative comparison in the appendix in Figure \ref{app:quality_11vs18}. In this figure, it is apparent, that while the 1.1m variant is struggling with general coherence, the 1.85m variant achieves better results in this regard. The notable improvements observed with the 1.85m parameter variant underscore its potential as a foundation for future research, encouraging further exploration into optimal configurations and scalability of the method.
Although a large UNet with 111 million parameters produces the best results, performance drops with fewer parameters. Reducing the number of parameters in UNet configurations leads to a noticeable degradation in FID and KID metrics. The smallest UNet configuration with 652k parameters records an FID of 175.3 and a KID of 0.142. In contrast, FourierDiff-NCA, with 526k parameters, achieves an FID of 60.96 and a KID of 0.031. FourierDiff-NCA's localized interaction architecture contributes to stable performance and improved parameter efficiency.

\begin{figure}[htbp]
 \centering
  {\includegraphics[width=.98\linewidth]{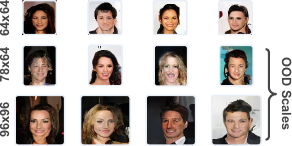}}
  \caption{Out-Of-Distribution image synthesis with FourierDiff-NCA (1.85m) of different scales and shapes.}
  \label{fig:OODImage}
\end{figure}

\subsection{Ablation}

The ablation study shown in Table \ref{tab:ablation} provides insights into the specific effects of the different hyperparameters of the FourierDiff-NCA model. The best results are achieved with our basic FourierDiff-NCA setup that uses 96 channels, a hidden size of 512, and 20 steps in the image space. Increasing the number of steps to $s=30$ negatively impacts the FID and KID values, as learning a meaningful rule becomes difficult. On the other hand, when $s=10$, the model does not have the perceptive range to incorporate enough local information to perform proper diffusion in the image space. Decreasing the hidden size to $h=256$ or the channel size to $c=48$ both has a negative effect on FID and KID, emphasizing the need for a sufficiently large number of parameters to capture the distribution of the underlying dataset. The impact of a smaller channel size is less severe than the reduction of the hidden size, even though FourierDiff-NCA with $h=256$ has twice as many parameters as the setup with $c=48$. The results show that careful calibration of these parameters can lead to optimal performance, and fine-tuning this balance will be a focus of future work.

\begin{table}[htbp]
\centering
\begin{tabular}{|c|c|c|c|}
    \hline
    \textbf{Setup} & \textbf{FID} $\downarrow$ & \textbf{KID} $\downarrow$ & \textbf{\# Param.} $\downarrow$ \\
    \hline
    \hline
    Basic & \textbf{43.86} & \textbf{0.022 $\pm$ 0.015} & 1,101,216 \\ 
    $s=30$ & 47.98 & \textbf{0.022 $\pm$ 0.015} & 1,101,216 \\ 
    $s=10$ & 61.47 & 0.029 $\pm$ 0.018 & 1,101,216 \\ 
    $h=256$ & 77.22 & 0.046 $\pm$ 0.025 & 737,184 \\ 
    $c=48$ & 60,96 & 0.031 $\pm$ 0.020 & \textbf{525,792} \\
    \hline
\end{tabular}
\caption{Ablation Results of FourierDiff-NCA, where $h = hidden size$, $c = channel size$ and $s = steps$. If not stated otherwise, $h=512$, $c=96$ and $s=20$.}
\label{tab:ablation}
\end{table}

\subsection{Out-Of-Distribution Scale: Image Synthesis}
Neural Cellular Automata have demonstrated the ability to generalize well beyond the specific training setup. Thus, we hypothesize that NCAs can handle larger image sizes and capture information beyond the pixel details of the input image through effective generalization. We \emph{synthesize images several times larger than the training size} to test this. The results, shown in Figure \ref{fig:OODImage}, clearly show the ability of the model to not only produce face-like images at scales up to $96\times96$ but also to add detail to the newly available space that is not present in the lower-resolution training images. This property is particularly evident in the finer details of the eyes and general facial features, which are more complex than observed at the training scale. As we expand the world's scale to $78\times64$, its content stretches while maintaining its relative orientation unchanged. Although some shortcomings in visual fidelity are apparent, it is likely that future improvements in the overall generation capabilities or OOD-specific optimizations of FourierDiff-NCA will improve OOD results.

\begin{figure}[htbp]
 \centering
  {\includegraphics[width=.98\linewidth]{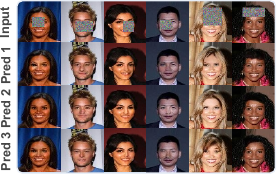}}
  \caption{FourierDiff-NCA (1.85m) can generate multiple predictions for a partially noisy image. The top row shows the input, while the other three rows show the predictions of the model.}
  \label{fig:Inpainting}
\end{figure}

\subsection{Inpainting}
We investigate the capabilities of FourierDiff-NCA in the inpainting task, where parts of the original image are intentionally obscured, and the model has to reconstruct the missing portions. This is done by placing a noisy square over an area of the image and then evaluating the model's predictions. Figure \ref{fig:Inpainting} shows several examples of the model's predictions given these modifications to the input. Although minor inaccuracies and occasional discolorations can be seen in some cases, the results generally confirm the model's ability to infer and approximate the hidden content. In comparison to UNet-based methods, FourierDiff-NCA has an edge, as the diffusion process in image space exclusively happens in the altered parts of the image. Thanks to the flexible model architecture, we can activate solely the pixels selected for inpainting, reducing the computational requirements of the selective diffusion to the reduced patch alone.

\begin{figure}[htbp]
 \centering
  {\includegraphics[width=.98\linewidth]{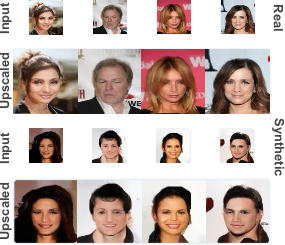}}
  \caption{FourierDiff-NCA (1.85m) captures sub-pixel information during training. Due to the unique model architecture, we can upscale images beyond the training size of $64 \times 64$ while adding additional details. In this case, we upscale $64\times64$ real and synthetic images to $128\times128$ images. }
  \label{fig:Supersample}
\end{figure}

\subsection{Image Upscaling}
Our investigation of FourierDiff-NCA showed that it can capture subpixel information and generate out-of-distribution (OOD) results. This intriguing result inspired us to explore the capacity of FourierDiff-NCA further. In this experiment, we investigate how well FourierDiff-NCA can upscale existing images to higher resolutions. 
To achieve our objective, we start with real and synthetic low-resolution images of size $64\times64$. These images are initially upscaled to $128\times128$ using a naive approach with nearest-neighbor interpolation. In this upscaling process, we essentially create a grid where the original pixels form the '1' positions in a pattern of [[1, 0], [0, 0]], leaving the '0' positions as newly added pixels. After this upscaling, we introduce 90\% noise and then employ FourierDiff-NCA for denoising. Importantly, in the backward diffusion process, we specifically target only the newly added pixels for updates. Additionally, during each update, we blend in 2\% of the original image data to these pixels, enhancing the integration of the upscaled and original image components. The results are illustrated in Figure \ref{fig:Supersample}. FourierDiff-NCA not only introduces new details in upscaled images but also integrates these with existing information, enhancing overall detail for a richer high-resolution output, affirming our original hypothesis that NCAs can generalize beyond the training size.

\begin{figure}[t]
 \centering
  {\includegraphics[width=.98\linewidth]{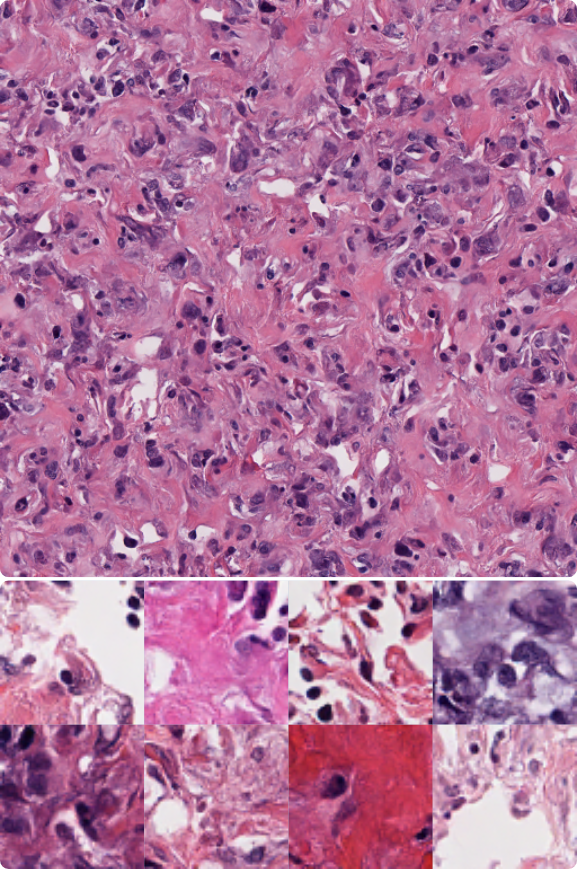}}
  \caption{Diff-NCA can train on images with a fixed resolution of $64\times64$ pixels and infer different resolutions, e.g., $512\times512$ pixels (top). The ability of Diff-NCA to generate different tissue types is shown in the bottom image, where eight generated samples are shown. }
  \label{fig:Gigapixel}
\end{figure}

\subsection{Fullscale Pathology: Image Synthesis}
A prevalent limitation of current diffusion and image generation models is that they are bound to fixed input sizes. This constraint becomes problematic when working with scenarios that require unconstrained world scales, such as digital pathology scans or satellite imagery. In such cases, global information is not essential to create an image that makes sense, as a coherent image can be synthesized by local communication alone. While UNet-based DDMs can generate individual patches, synthesizing a seamless, continuous image that appears unified is beyond the capabilities of current architectures.

Diff-NCA is designed to address these local constraints, allowing the generation of image patches of theoretically unlimited size while ensuring a seamless result. This is ensured by choosing a step size small enough that the model merely diffuses based on local features. In this process, local communication preserves the continuity of the image so that all components are aligned without disjointness or abrupt transitions. In Figure \ref{fig:Gigapixel}, we illustrate this by synthesizing a $512 \times 512$ digital pathology scan, demonstrating the model's ability to produce a seamless and visually coherent result. A key difference between increasing the world scale of Diff-NCA and FourierDiff-NCA after training is that Diff-NCA produces more tissue of the same scale, while FourierDiff-NCA increases its detail since it has been trained on integrating the global information. Each of these methods holds its distinct utility and application. Diff-NCA's ability to recognize and synthesize different tissue types and stains highlights its versatile and promising potential in unconstrained imaging tasks.

\section{Conclusion}
We introduce Diff-NCA and FourierDiff-NCA as a jumpstart for Neural Cellular Automata-based denoising methods. Diff-NCA and FourierDiff-NCA are designed to produce high-quality images with merely 331k and 1.1m parameter counts, respectively. Diff-NCA is strategically optimized to utilize local features in specific applications such as pathology where they are sufficient. As an advancement of Diff-NCA, FourierDiff-NCA addresses the complex challenge of global communication within NCA by using the Fourier space for instantaneous global communication that is subsequently refined in image space. The effectiveness of these proposed architectures is demonstrated by generating $512 \times 512$ pathology patches with Diff-NCA and $64\times64$ CelebA images with FourierDiff-NCA. FourierDiff-NCA outperforms the NCA-based image generation model VNCA and four times bigger UNet-based DDMs. FourierDiff-NCA can perform super-resolution, out-of-distribution image generation, and inpainting without additional training. These results highlight the potential of Diff-NCA and FourierDiff-NCA as compelling alternative methods for efficient and versatile imaging that offer giga-pixel image generation in low-resource environments, democratizing generative learning.

\section*{Impact Statement}
In this paper, we introduce Diff-NCA and FourierDiff-NCA, Neural Cellular Automata-based DDM architectures that significantly reduce parameter count in image synthesis, compared to conventional UNet architectures. This reduction presents a notable advancement in model compactness, potentially leading to broader applications, especially in environments with limited computational resources.
\bibliography{paper}
\bibliographystyle{icml2024}

\newpage
\phantom{todo}
\newpage
\appendix
\section{Neural Cellular Automata (NCA): Introduction}
\label{app:NCAIntro}
\begin{figure}[htbp]
 \centering
  {\includegraphics[width=.97\linewidth]{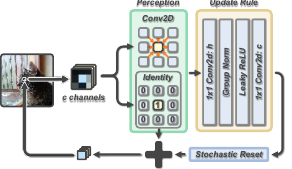}}
  \caption{An update step of a single cell.}
  \label{fig:singleCellUpdate}
\end{figure}

Neural Cellular Automata (NCA) represent a type of minimal model that deviates from traditional deep learning architectures by \emph{focusing on individual pixels or cells} within an image. This departure from a holistic view of the entire scene contributes to their compactness. This means that a single NCA looks only at one cell. To take into account global information, this same rule is replicated across all pixels of an image. Recognizing the need for information exchange beyond their individual states, these models are fed by an input layer-often in the form of a 3x3 convolutional layer to establish cellular communication with neighboring cells. A single update step of one cell is illustrated in Figure \ref{fig:singleCellUpdate}. Consequently, each cell acquires awareness of its own state while engaging in controlled interactions with its immediate environment. To handle more complicated and multi-faceted tasks, the NCA approach requires an iterative execution that allows the model to operate over multiple temporal iterations. This iterative approach facilitates the sharing of information across the entirety of the image, leveraging the ability to successfully perform complex operations orchestrated on a broader spatial scale.

In summary, Neural Cellular Automata represent an efficient architecture in which updates occur at the level of individual cells that are equipped with neighborhood cell communication. This approach, combined with the iterative application of said update rule, enables these models to handle complicated image analysis problems by orchestrating a coordinated flow of information across the entire image.

\subsection{Important concepts}
When considering the technical implementation of Neural Cellular Automata (NCAs), several key concepts need to be explored. First and foremost is the \textbf{step count}, which indicates the number of times the model is applied at the image scale. This parameter largely determines the perceptive range of a single NCA model and limits the communicative range. The perceptive range is further influenced by the \textbf{fire rate} that controls the stochastic cell updates, implemented by a stochastic reset mechanism. Furthermore, the firing rate serves as a critical factor in shaping a robust update rule within the model and enhances its ability to develop a more robust and adaptive rule.

\subsubsection{VRAM requirements}
Despite their lightweight architecture, training NCAs can impose significant demands on VRAM as the learned update rule is replicated across the whole image scale. Combined with the iterative model execution, this leads to an exponential increase in memory requirements highly dependent on image size. As all states have to be kept in the VRAM to perform backpropagation via gradient decent, a naive approach to running an NCA on a $256\times256$ image for only 60 steps results in VRAM usage of over 20 GB \cite{kalkhof2023med}.

\begin{figure}[htbp]
 \centering
  {\includegraphics[width=.97\linewidth]{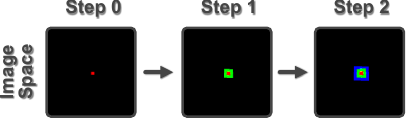}}
  \caption{Increase of the perceptive range of the red cell in image space.}
  \label{fig:perceptiveRange}
\end{figure}

\subsubsection{Perceptive Range / Communication}
The perceptive range of an NCA increases with its number of steps. In a naive setup, which uses a 3x3 convolution as the input to the model, each NCA can only communicate with its direct neighbors. This means propagating information across a 100x100 image subsequentially requires 100 steps. This makes it difficult and slow to acquire global information. Further, it has to be carefully balanced, as a perceptive range too big, makes it difficult for the model to learn a meaningful rule, while one that is too small, prevents the model from gaining global insight. It is further influenced by the fire rate, as a stochastic reset implies that this cell won't receive any updates in this timestep.

\subsubsection{Fire Rate}
In the context of NCAs, a common concept is the introduction of a firing rate that controls cellular updates. This rate, which is typically 50\%, determines whether the state of a cell changes in each iteration. The fire rate is implemented by canceling the cellular update for the corresponding step, by applying a stochastic reset. Consequently, the model is forced to adapt to the stochastic flow of information, which increases its robustness, as shown in the growth of cells from singular pixels \cite{mordvintsev2020growing}.

\subsubsection{Channels}
The channels used include both input data, such as the image or in our case the noisy image, and additional channels that are used for storing information inbetween the NCAs update steps. Ensuring adequate channel size is critical, as an insufficient capacity can compromise the ability of the NCA to store important information between steps.

\section{Fourier Communication}
\label{app:fourierCom}
\begin{figure}[htbp]
 \centering
  {\includegraphics[width=.97\linewidth]{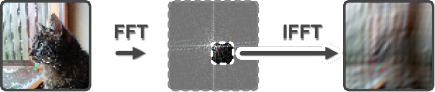}}
  \caption{Cutting away 93.75\% of the information in Fourier space drastically reduces the size, yet important low-frequency information is preserved.}
  \label{fig:fft_reasoning}
\end{figure}

The Fourier space is characterized by a natural arrangement in which the low-frequency information is concentrated in the center of the space. Notably, a significant portion of the important low-frequency data is preserved even when 93.75\% of the information is removed, as illustrated in Figure \ref{fig:fft_reasoning}. This persistence of essential low-frequency content has sufficient information value to allow orientation in a global context. The unique property of Fourier space to retain key information despite significant data reduction underscores its suitability for retaining fundamental spatial features across scales.

\begin{figure}[htbp]
 \centering
  {\includegraphics[width=.97\linewidth]{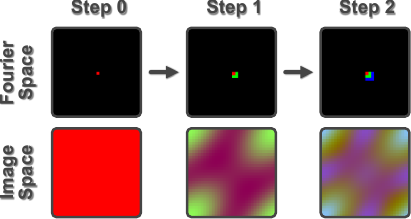}}
  \caption{Simplified visualization, of how communication between NCA's flows in the Fourier space and how it translates to the image space. A single step is enough to communicate across the whole image, whereas extra steps refine the signal sent.}
  \label{fig:fft_Steps}
\end{figure}

Taking advantage of its inherent structural features, the Fourier space turns out to be a strategically advantageous channel for accelerating global communication within local interaction models. Said process involves the NCA performing a particular number of steps in Fourier space, leading to the accumulation of comprehensive global knowledge. Following this phase, the accumulated knowledge is transferred into the image space through an inverse Fast Fourier Transform. Figure \ref{fig:fft_Steps} provides a simplified representation of this information flow, illustrating the interplay between Fourier space and image space. This novel use of the characteristics of Fourier space highlights its central role in enabling efficient global communication of locally interacting NCAs.

\section{Qualitative Comparison FourierDiff-NCAs}
\begin{figure}[htbp]
 \centering
  {\includegraphics[width=.97\linewidth]{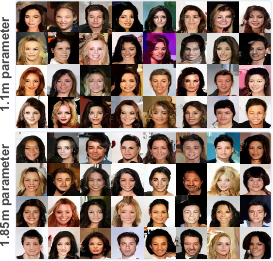}}
  \caption{Qualitative comparison between FourierDiff-NCA 1.1m and FourierDiff-NCA 1.85m.}
  \label{app:quality_11vs18}
\end{figure}

A qualitative comparison of FourierDiff-NCA models with 1.1m and 1.85m parameters is presented in Figure \ref{app:quality_11vs18}. Despite a modest 3\% increase in FID, the KID score sees a substantial 28\% improvement, reflecting the enhanced coherence of the generated images.

\section{Quantitative Results}

The precise quantitative results comparing FourierDiff-NCA, VNCA, and the UNets (S, M, B, L) compared in our figure can be found in Table \ref{tab:quant}.

\begin{table}[htbp]
\centering
\begin{tabular}{|l|c|c|c|}
    \hline
    \textbf{Method} & \textbf{FID} $\downarrow$ & \textbf{KID} $\downarrow$ & \textbf{\# Param.} $\downarrow$ \\
    \hline
    \hline
    FD-NCA (1.85m) & \textbf{42.69} & \textbf{0.016 $\pm$ 0.012} & 1,849,632 \\ 
    FD-NCA & 43.86 & 0.022 $\pm$ 0.015 & 1,101,216 \\ 
    VNCA & 299.99 & 0.338 $\pm$ 0.032 & 9,732,416\\
    UNet S & 175.3 & 0.142 $\pm$ 0.009 & 652,195 \\
    UNet M & 128.2 & 0.089 $\pm$ 0.009 & 3,940,547 \\
    \hline
    \hline
    UNet B & 20.1 & 0.005 $\pm$ 0.002 & 17,217,923 \\
    UNet L & 18.2 & 0.004 $\pm$ 0.002 & 111,051,779 \\ 
    \hline
    
\end{tabular}
\caption{Quantiative comparison of FourierDiff-NCA (here FD-NCA) and UNet-based DDMs.}
\label{tab:quant}
\end{table}

\section{Experimental Setup}
To ensure better reproducibility, detailed experiment settings are given below. All experiments performed are done with PyTorch \cite{paszke2019pytorch}, with comprehensive descriptions complementing the full codebase, which will be released after acceptance.

\begin{table*}[htbp]
\centering
\begin{tabular}{|l|c|c|c|c|}
    \hline
    \textbf{UNet} & \textbf{Channel dimension} & \textbf{Resnet block per resolution} & \textbf{Attention resolution} $\downarrow$ & \textbf{\# Param.} \\
    \hline
    \hline
    Small & (32; 64) &  1 & (-) & 652,195 \\
    Medium & (32; 64; 128) &  2 & (16) & 3,940,547 \\
    Big & 	(64; 128; 128; 256) &  2 & (16) & 17,217,923 \\
    Large &  (128; 128; 256; 256; 512; 512) & 2 & (16; 8) & 111,051,779 \\
    \hline
    
\end{tabular}
\caption{Configuration of UNets.}
\label{tab:UNets}
\end{table*}

\subsection{Diff-NCA and FourierDiff-NCA}
All conducted experiments utilize the Adam optimizer \cite{kingma2014adam}. The chosen hyperparameters include a learning rate of $1.6 \times 10^{-3}$, a learning rate gamma of $0.9999$, betas for the learning rate as $(0.9, 0.99)$, and epsilon ($\epsilon$) set at $1 \times 10^{-8}$. The models undergo training for 200,000 steps, utilizing a batch size of 16. Detailed configurations for FourierDiff-NCA are outlined in Listing \ref{lst:FourierDiff}, while those for Diff-NCA are provided in Listing \ref{lst:Diff}.
\\
\begin{lstlisting} [caption={Configuration of FourierDiff-NCA},captionpos=b, label={lst:FourierDiff}, breaklines=true]
    FourierDiff_NCA(
      (norm_four): GroupNorm(1, 512, eps=1e-05, affine=True)
      (perceive_four): Conv2d(196, 192, kernel_size=(3, 3), stride=(1, 1), padding=(1, 1))
      (fc0_four): Conv2d(388, 512, kernel_size=(1, 1), stride=(1, 1))
      (fc1_four): Conv2d(512, 192, kernel_size=(1, 1), stride=(1, 1))
      (mulCond0_four): Sequential(
        (0): Conv2d(4, 128, kernel_size=(1, 1), stride=(1, 1))
        (1): SiLU()
        (2): Conv2d(128, 388, kernel_size=(1, 1), stride=(1, 1))
      )
      (mulCond1_four): Sequential(
        (0): Conv2d(4, 128, kernel_size=(1, 1), stride=(1, 1))
        (1): SiLU()
        (2): Conv2d(128, 512, kernel_size=(1, 1), stride=(1, 1))
      )
      (embedding_four): Sequential(
        (0): Conv2d(16, 256, kernel_size=(1, 1), stride=(1, 1))
        (1): SiLU()
        (2): Conv2d(256, 4, kernel_size=(1, 1), stride=(1, 1))
      )
      (norm_img): GroupNorm(1, 512, eps=1e-05, affine=True)
      (perceive_img): Conv2d(100, 96, kernel_size=(3, 3), stride=(1, 1), padding=(1, 1), padding_mode=reflect)
      (fc0_img): Conv2d(196, 512, kernel_size=(1, 1), stride=(1, 1))
      (fc1_img): Conv2d(512, 96, kernel_size=(1, 1), stride=(1, 1))
      (mulCond0_img): Sequential(
        (0): Conv2d(4, 128, kernel_size=(1, 1), stride=(1, 1))
        (1): SiLU()
        (2): Conv2d(128, 196, kernel_size=(1, 1), stride=(1, 1))
      )
      (mulCond1_img): Sequential(
        (0): Conv2d(4, 128, kernel_size=(1, 1), stride=(1, 1))
        (1): SiLU()
        (2): Conv2d(128, 512, kernel_size=(1, 1), stride=(1, 1))
      )
      (embedding_img): Sequential(
        (0): Conv2d(16, 256, kernel_size=(1, 1), stride=(1, 1))
        (1): SiLU()
        (2): Conv2d(256, 4, kernel_size=(1, 1), stride=(1, 1))
      )
    )
\end{lstlisting}

\begin{lstlisting} [caption={Configuration of Diff-NCA},captionpos=b, label={lst:Diff}, breaklines=true]

Diff_NCA(
  (norm_img): GroupNorm(1, 512, eps=1e-05, affine=True)
  (perceive_img): Conv2d(98, 96, kernel_size=(3, 3), stride=(1, 1), padding=(1, 1), padding_mode=reflect)
  (fc0_img): Conv2d(194, 512, kernel_size=(1, 1), stride=(1, 1))
  (fc1_img): Conv2d(512, 96, kernel_size=(1, 1), stride=(1, 1))
  (mulCond0_img): Sequential(
    (0): Conv2d(2, 128, kernel_size=(1, 1), stride=(1, 1))
    (1): SiLU()
    (2): Conv2d(128, 194, kernel_size=(1, 1), stride=(1, 1))
  )
  (mulCond1_img): Sequential(
    (0): Conv2d(2, 128, kernel_size=(1, 1), stride=(1, 1))
    (1): SiLU()
    (2): Conv2d(128, 512, kernel_size=(1, 1), stride=(1, 1))
  )
  (embedding_img): Sequential(
    (0): Conv2d(4, 256, kernel_size=(1, 1), stride=(1, 1))
    (1): SiLU()
    (2): Conv2d(256, 2, kernel_size=(1, 1), stride=(1, 1))
  )
)
\end{lstlisting}

\subsection{VNCA}
We use the official CelebA specific implementation of VNCA \cite{palm2022variational} which can be found at \url{https://github.com/rasmusbergpalm/vnca}.

\subsection{UNets}
The training of all UNets employs the Adam optimizer with a learning rate of $3 \times 10^{-5}$, betas set at $(0.5, 0.999)$, and epsilon ($\epsilon$) at $1 \times 10^{-6}$. A training duration of 200,000 steps is undertaken using a batch size of 32. Refer to Table \ref{tab:UNets} for detailed layer configurations of each UNet model.



\end{document}